\documentclass[letterpaper, 10 pt, conference]{ieeeconf}  
\IEEEoverridecommandlockouts                              
\usepackage{tikz}
\usetikzlibrary{shapes,arrows,calc}

\tikzstyle{decision} = [diamond, draw, fill=gray!20,
    text width=4.5em, text badly centered, node distance=2.5cm, inner sep=0pt]
\tikzstyle{block} = [rectangle, draw, fill=gray!20,
    text width=5em, text centered, rounded corners, minimum height=4em]
\tikzstyle{line} = [draw, very thick, color=black!50, -latex']
\tikzstyle{cloud} = [draw, ellipse,fill=red!20, node distance=2.5cm,
    minimum height=2em]

\newcommand{\norm}[1]{\lVert#1\rVert}

\usepackage{graphics} 
\usepackage{etex}
 \usepackage{booktabs}
\usepackage{amsmath,bm} 
\usepackage{amssymb}  
\usepackage{stfloats}
\usepackage{multirow}
\usepackage{subfigure}
\usepackage{algorithm}
\usepackage[noend]{algorithmic}
\usepackage[pdfborder=0]{hyperref}
\usepackage{float}
\usepackage[usenames,dvipsnames]{pstricks}
\usepackage{epsfig}
\usepackage{pst-grad} 
\usepackage{pst-plot} 

\title{\LARGE \bf
An Eigenshapes Approach to Compressed Signed Distance Fields and Their Utility in Robot Mapping
}

\author{%
    \authorblockN{ Daniel R. Canelhas, Erik Schaffernicht, Todor Stoyanov, Achim J. Lilienthal}
    \authorblockA{
        Center of Applied Autonomous Sensor Systems (AASS), {\"O}rebro University, Sweden
    }
    \authorblockN{Andrew J. Davison}
    \authorblockA{Department of Computing, Imperial College London, UK}
}

\begin{document}

\maketitle
\thispagestyle{empty}
\pagestyle{empty}

\begin{abstract}
In order to deal with the scaling problem of volumetric map representations we propose spatially local methods for high-ratio compression of 3D maps, represented as truncated signed distance fields. We show that these compressed maps can be used as meaningful descriptors for selective decompression in scenarios relevant to robotic applications. As compression methods, we compare using PCA-derived low-dimensional bases to non-linear auto-encoder networks and novel mixed architectures that combine both. Selecting two application-oriented performance metrics, we evaluate the impact of different compression rates on reconstruction fidelity as well as to the task of map-aided ego-motion estimation. It is demonstrated that lossily compressed distance fields used as cost functions for ego-motion estimation, can outperform their uncompressed counterparts in challenging scenarios from standard RGB-D data-sets. 
\end{abstract}

\section{Introduction}
A signed distance field (SDF), sometimes referred to as a distance function, is an implicit surface representation that embeds geometry into a scalar field whose defining property is that its value represents the distance to the \textit{nearest} surface of the embedded geometry. Additionally, the  field is positive outside the geometry, i.e., in free space, and negative inside. 
SDF's have been extensively applied to e.g.  speeding up image-alignment \cite{fitzgibbon2003robust} and raycasting \cite{hart1996sphere} operations as well as collision detection \cite{fuhrmann2003distance}, motion planning \cite{hoff1999fast} and articulated-body motion tracking \cite{schmidt2014dart}. The truncated SDF \cite{curless1996volumetric} (TSDF), which is the focus of the present work, side-steps some of  the difficulties that arise when fields are computed and updated based on incomplete information. This has proved useful in applications of particular relevance to the field of robotics research: accurate scene reconstruction (\cite{newcombe2011kinectfusion}, \cite{whelan2012kintinuous}, \cite{roth2012moving}) as well as for rigid-body (\cite{canelhas2013sdf}, \cite{bylow2013sdf}) pose estimation. 
 \par
The demonstrated practicality of distance fields and other voxel-based representations such as occupancy grids\cite{ elfes1989occupancy} and the direct applicability of a vast arsenal of image processing methods to such representations make them a compelling research topic. However, a major drawback in such representations is the large memory requirement for storage which severely limits their applicability for large-scale environments. For example, a space measuring $20\times20\times4 m^3$ mapped with voxels of 2cm size requires at least 800MB at 32 bits per voxel. 
\par
Mitigating strategies such as cyclic buffers (\cite{whelan2012kintinuous}, \cite{roth2012moving}), octrees (\cite{frisken2000adaptively}, \cite{zeng2012memory}), and key-block swapping \cite{newcombe2014phd}, have been proposed to limit the memory cost of using volumetric distance-fields in very different ways.     
In the present work, we address the issue of volumetric voxel-based map compression by an alternative strategy. We propose encoding (and subsequently decoding) the TSDF in a low-dimensional feature space by projection onto a learned set of basis (eigen-) vectors derived via principal component analysis \cite{wold1987principal} (PCA) of a large data-set of sample reconstructions. We also show  that this compression method preserves important structures in the data while filtering out noise, allowing for more stable camera-tracking to be done against the model, using the SDF Tracker \cite{canelhas2013sdf}  algorithm. We show that this method compares favourably to non-linear methods based on auto-encoders (AE) in terms of compression, but slightly less so in terms of tracking performance. Lastly, we investigate whether combinations of PCA-based and AE strategies in mixed architectures provide better maps than either system on its own but find no experimental evidence to support this.

The proposed compression strategies can be applied to scenarios in which robotic agents with limited on-board memory and computational resources download the maps from sensor-enabled work environments. In this context, the low dimensional features produced by the compression method serve as descriptors, providing an opportunity for the robot to, still in the descriptor-space, make the decision to selectively decompress regions of the map that may be of particular interest. A proof of concept for this scenario is presented in Sec. \ref{sec:results}.
\par
The remainder of the paper is organized as follows:
An overview on related work in given in section \ref{sec:related}. In section \ref{sec:prelim} we formalize the definition of TSDF's, and present a very brief introduction to the topics of PCA and AE networks. In section \ref{sec:method} we elaborate on the training data used, followed by a description of our evaluation methodology. Section \ref{sec:results} contains experimental results, followed by section \ref{sec:conclusions} with our conclusions and lastly, some possible extensions to the present work are suggested in section \ref{sec:future}.
\section{Related Work}\label{sec:related}
Our work is perhaps most closely related to sparse coded surface models \cite{ruhnke2013compact} which use $k$-SVD \cite{aharon2006svd} (a linear projection method)  to reduce the dimensionality of textured surface patches. Another recent contribution in this category is the Active Patch Model for 2D images \cite{mao2014active}. Active patches consist of a dictionary of data patches in input space that can be warped to fit new data. A low-dimensional representation is derived by optimizing the selection of patches and pre-defined warps that best use the patches to reconstruct the input. The operation on surface patches instead of volumetric image data is more efficient for compression for smooth surfaces, but may require an unbounded number of patches for arbitrarily complex geometry. As an analogy, our work can be thought of as an application of Eigenfaces \cite{turk1991eigenfaces} to the problem of 3D shape compression and low-level scene understanding.
Operating directly on a volumetric representation, as we propose, has the advantage of a constant compression ratio per unit volume, regardless of the surface complexity, as well as avoiding the problem of estimating the optimal placement of patches. 
The direct compression of  the TSDF also permits the proposed method to be integrated into several popular algorithms that rely on this representation, with minimal overhead. 
There are a number of data-compression algorithms designed for directly compressing volumetric data. Among these we find video and volumetric image compression (\cite{richardson2004h},\cite{marcellin2002jpeg2000}), including work dealing with distance fields specifically \cite{jones2004distance}. Although these methods produce high-quality compression results, they typically require many sequential operations and complex extrapolation and/or interpolation schemes. A side-effect of this is that these compressed representations may require information far from the location that is to be decoded. They also do not generate a mapping to a  feature space wherein similar inputs map to similar features so possible uses as descriptors are limited at best.
\section{Preliminaries}\label{sec:prelim}
\subsection{Truncated Signed Distance Fields (TSDF)}
TSDFs are 3-dimensional image structures that implicitly represent geometry by sampling, typically on a uniform lattice, the distance to the nearest surface. A sign is used to indicate whether the distance is sampled from within a solid shape (negative, by convention) or in free space (positive).  The approximate location of surfaces can be extracted as the zero level set. Let,
\begin{equation}
d'(\bm{p})  : \mathbb{R}^3 \rightarrow \mathbb{R}
\end{equation}
be defined as the distance field of some arbitrary closed surface in $\bm{Q}$ in $\mathbb{R}^3$ ,
\begin{equation}
d'(\bm{p})  = \mathop{argmin}_{\bm{q}\in \bm{Q}} \norm{\bm{p}-\bm{q}}_2. 
\end{equation}
Given the closed (no holes) property of the surface, one may assume that every surface point has an associated outward-oriented normal vector $ \bm{n}(\bm{q})$. The expression $\mathbb{I_{\pm}}(p)=sign( \bm{n}(\bm{q})^T \cdot( \bm{p}-\bm{q}) )$, then consistently attributes a sign to indicate on which side of the surface $\bm{p}$ is located. Finally, truncating the value of the field in an interval  $\left[ d_{min}, d_{max} \right]$ produces the TSDF, 
\begin{equation}
d(\bm{p})  : \mathbb{R}^3 \rightarrow \left[ d_{min}, d_{max} \right]
\end{equation}
defined, for any closed surface, as
\begin{equation}
d(\bm{p})  =  min(d_{max,}max(d_{min},\mathbb{I_{\pm}}(p)\mathop{argmin}_{\bm{q}\in \bm{Q}} \norm{\bm{p}-\bm{q}}_2)). 
\end{equation}
\subsection{Principal Component Analysis (PCA)}
PCA \cite{wold1987principal} is a method for obtaining a linear transformation into a new  orthogonal coordinate system. In this system, the first dimension is associated with the direction, in the data, that exhibits the largest variance. The second dimension is aligned with a direction, perpendicular to the first, along which the second most variance is exhibited and so on.  We achieve this by the common method of applying a singular value decomposition (SVD) to the data matrix after subtracting the mean from each sample. 
Since PCA encoding, applied to non-centred data, needs to store the mean of the input for later decoding steps, we extract \textit{31, 63} and \textit{127} components and use one additional slot to store the mean, resulting in compact representations of \textit{32, 64} and \textit{128} elements. 
\subsection{Artificial Neural Network}
Training an artificial neural network (ANN)  as an auto-encoder \cite{rumelhart1988learning} can be done in a straightforward manner by setting its desired output to be equal to its input and employing an optimization method of choice to minimize the resulting error. For some form of encoding to occur, it is required that somewhere in between the input layer and output layer, there exists an intermediary hidden layer whose output is of smaller dimension than the input (or output). We refer to this intermediate "bottleneck" layer as a code or feature layer. 
The portion of the ANN up until the feature layer can then be treated as an encoder and the portion after is treated as a decoder. For practical reasons (particularly when layer-wise unsupervised pre-training is involved \cite{hinton2006reducing}) it makes sense to keep the encoder and decoder symmetric. 
\section{Methodology}\label{sec:method}
\subsection{Training data }
\begin{figure}[tbp]
\begin{center}
\includegraphics[width =1\linewidth]{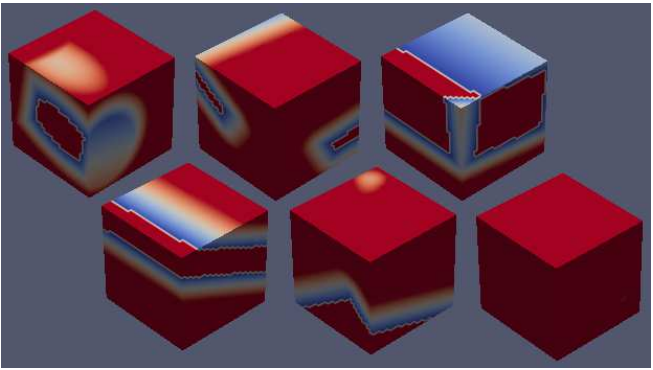}
\caption{Examples from the synthetic data-set showing a variety of shapes represented by truncated distance fields, sampled onto a small volume containing 4096 voxels. }
\label{fig:synthetic_data}
\vspace{-0.8cm}
\end{center}
\end{figure}

The data-set used for training is a collection of synthetic TSDFs, procedurally generated using \textbf{libsdf} \footnote{https://bitbucket.org/danielcanelhas/libsdf}, an open-source C++ library that implements simple implicit geometric primitives (as described in \cite{hart1996sphere}, \cite{distfun}). The library was used to randomly pick a sequence of randomly parametrized shapes from several shape categories.
 A random displacement and rotation is applied to each shape and  the distance field is sampled (truncated to $d_{min } = -0.04$ and $d_{max}=0.1$) into a cubic lattice of $16 \times1 6\times 16$ voxels. Some examples from our synthetic data-set can be seen in Fig. \ref{fig:synthetic_data}.
 \par
 We note that planes, convex edges and corners can be extracted as parts of cuboids, thus we consider such shapes as special cases of the category describing cuboids since sampling volume tends to capture only parts of the whole shape, often resulting in planar, edge or corner fragments. By a similar line of reasoning, we employ a parametric barrel-like shape to model curved convex edges and cylinders. The final shape category used in the data-set is a concave corner shape (representing 2-plane concave edges as a special case). When considering only surface, without orientation, a convex corner is indistinguishable from a concave one, but since we are interested in signed distance fields, the orientation matters.
The use of synthetic data allows generating training examples in a vast number of poses, with a greater degree of geometric variation than would be feasible to collect manually through scene reconstructions alone.  
 \par
\begin{figure}[tbp]
\begin{center}
\includegraphics[width =0.6\linewidth]{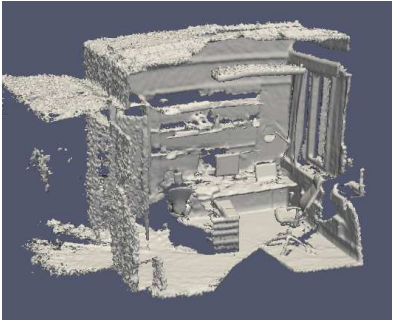}
\caption{Examples from the real-world data, showing the extracted zero level set as a polygonal surface mesh. The picture depicts a partial reconstruction of a small office environment.}
\label{fig:real_data}
\vspace{-0.5cm}
\end{center}
\end{figure}

However, to add additional complexity beyond simple geometric primitives, the data-set is complemented with sub-volumes sampled from 3D reconstructions of real-world industrial and office environments c.f. Fig. \ref{fig:real_data}. These reconstructions are obtained by fusing sequences of depth images into a TSDF as described in \cite{curless1996volumetric}, given accurately estimated camera poses by the SDF Tracker algorithm (though any method with low drift would do just as well). 
\par
The sub-volumes are sampled  by taking $16 \times 16\times 16$  samples at every 8 voxels along each spatial dimension and permuting the indexing order along each dimension for every samples to generate 5 additional reflections at each location. Distance values are then mapped from the interval $\left[d_{min}, d_{max}\right]$ to $\left[0, 1\right]$ and saved. Furthermore, to avoid an uncontrolled amount of effort spent on learning models of empty space, sub-volumes for which the mean sum of normalized ($ \in\left[0, 1\right]$) distances is below $0.85$ are discarded, and a small proportion of empty samples is intentionally included instead. Defining our input dimension as $n = 4096$, with $m=200000$ samples, our data-set is then  $\bm{X} \in \lbrace\mathbb{R}^{m \times n} |0\leq x_{i,j} \leq 1 \rbrace$.
\subsection{Encoder Architecture}
Although the main focus of this paper is on a simple method: projection onto a basis of eigenvectors (principal components) of a large set of sampled reconstructions, alternatively using auto-encoder networks for dimensionalty reduction, we present and test a couple of extensions to these ideas combining both methods. 

\subsubsection{Parallel Encoding/Decoding}
\begin{figure}[h]
\begin{centering}
\includegraphics[width =0.75\linewidth]{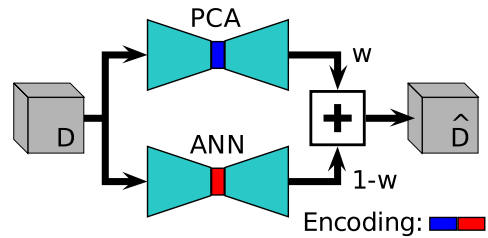}
\caption{Coupling the PCA and ANN encoder/decoder pairs in a parallel manner. Both are trained on the original data-set and their outputs are combined through a weighted sum.}
\label{fig:parallel}
\vspace{-0.25cm}
\end{centering}
\end{figure}

The first extension is a method to combine different encoders/decoders, inspired by ensemble theory \cite{dietterich2000ensemble} which states that classifiers in a committee perform better than any single classifier, if the individual decision making is independent of each other and better than chance. Applied to this problem, we propose to combine a PCA-based encoder with an ANN, as shown in Fig. \ref{fig:parallel}. 
For compression, the TSDF is encoded separately by both encoders. The allotment of code elements is split in some pre-determined way between the encoders. We use codes with total length of \textit{128} elements, for our experiments. The final code is simply the concatenation of both individual codes (shown as blue and red in the figures). Decoding is done independently by each decoder on their respective part of the code, and their outputs are added with weights $w \in \left[0, 1\right]$ and $1-w$. To provide the best-case performance of this approach, $w$ is computed by an approximate line-search, minimizing the reconstruction error. In practice, the cost of searching for an optimal $w$ for each encoded block may be prohibitive and real-time applications may instead favour a fixed weight for the entire map.
\subsubsection{Sequential Encoding/Decoding}
\begin{figure}[t!]
\begin{center}
\subfigure[] {
        \includegraphics[width =.28\linewidth]{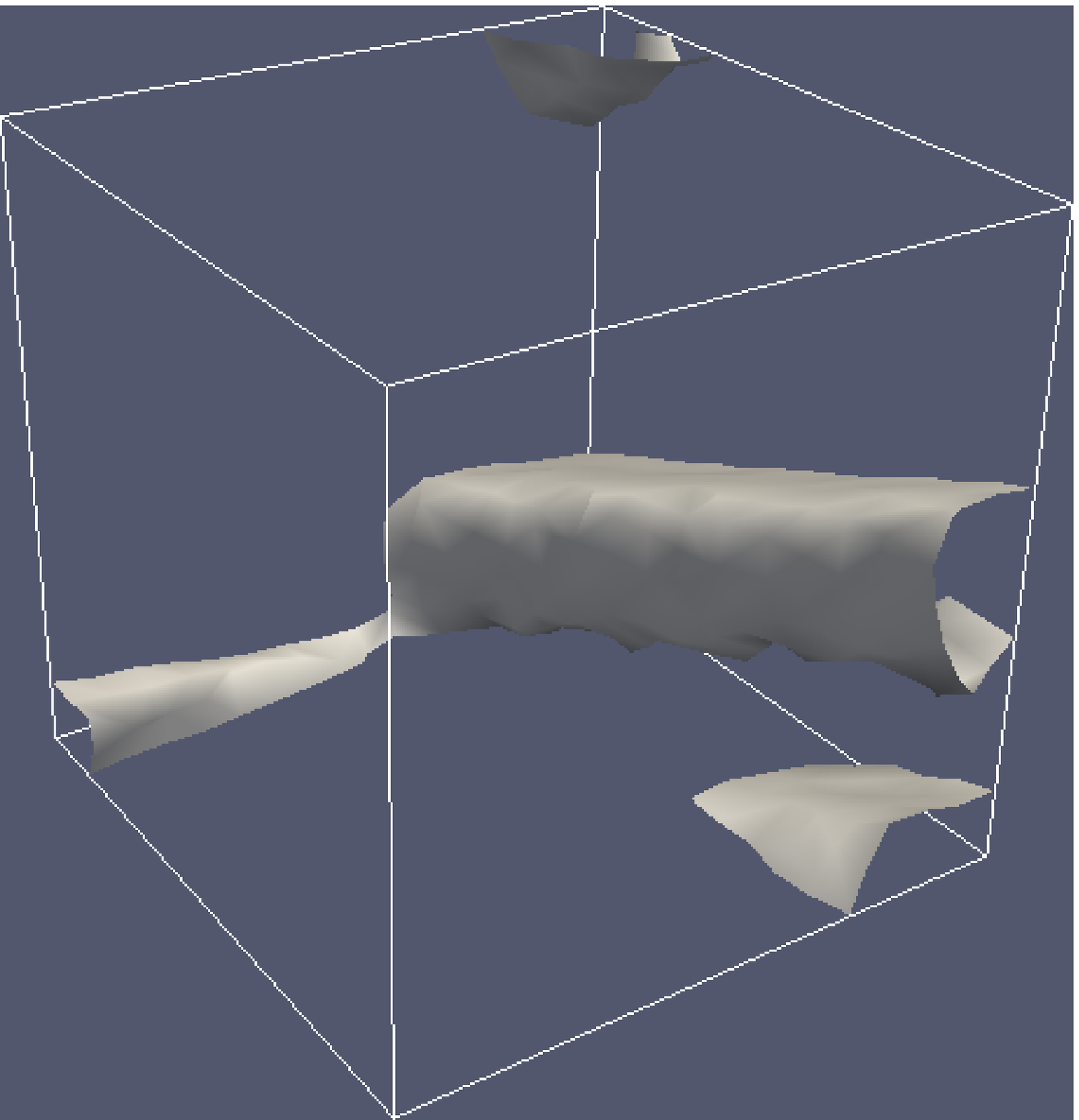}
	    \label{fig:seq_in}
}
\subfigure[] {
        \includegraphics[width =.28\linewidth]{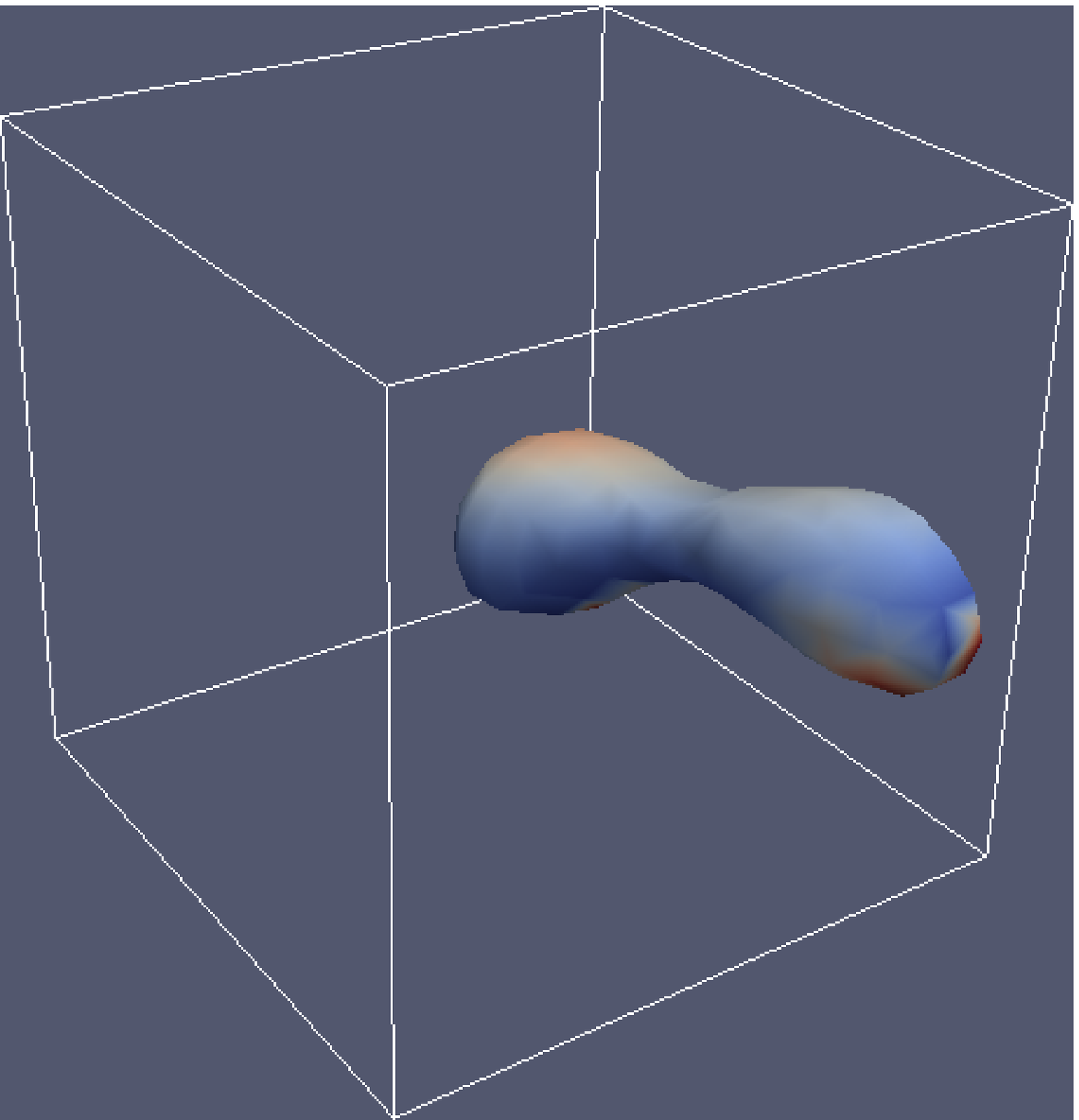}
	    \label{fig:seq_pca}
}
\subfigure[] {
        \includegraphics[width =.28\linewidth]{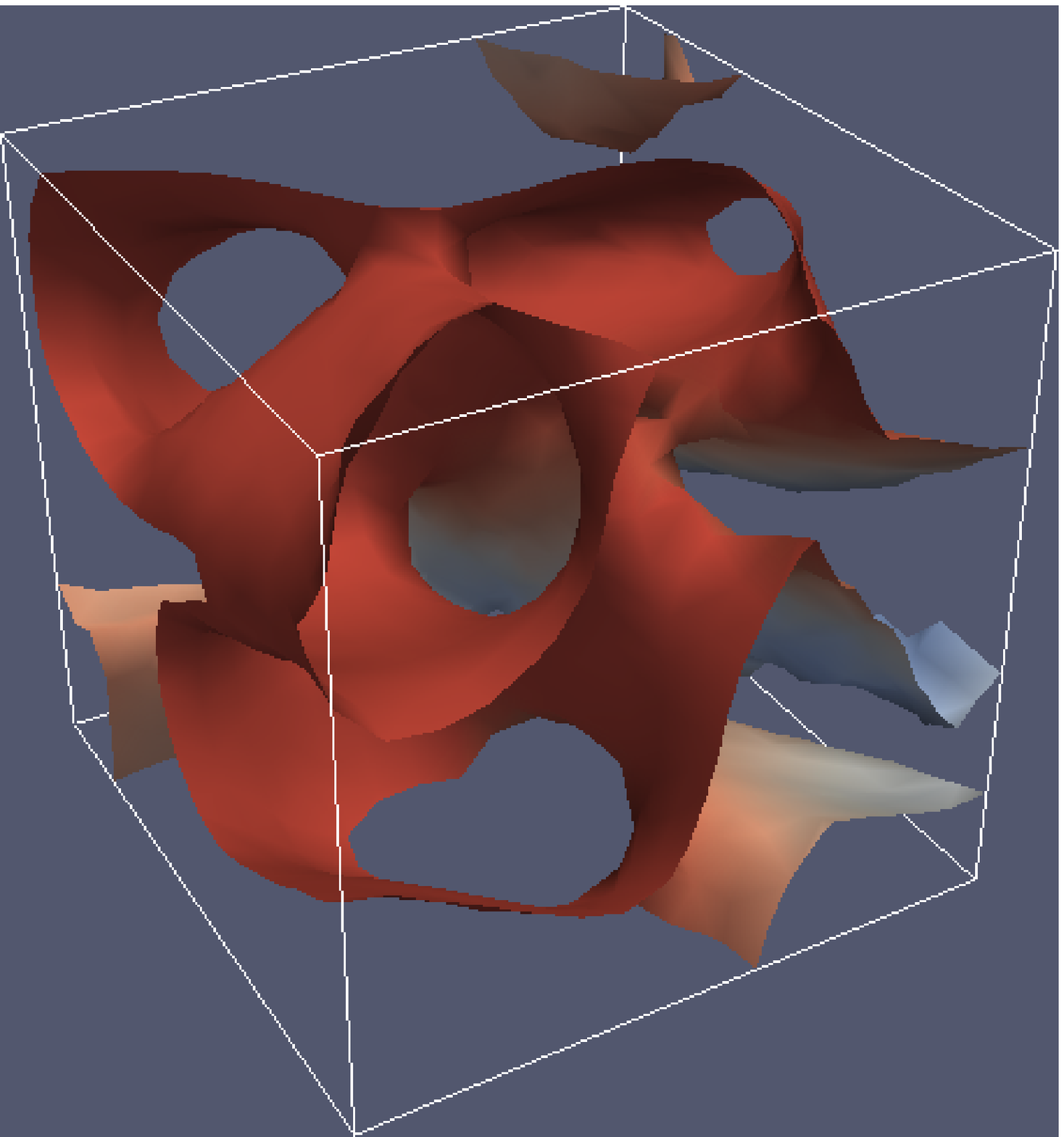}
	    \label{fig:seq_res}
}
\vspace{-0.25cm}
\caption{ The residual volume contains more complex data, but is evidently not a random signal. Input data is shown as an extracted surface in \subref{fig:seq_in}. PCA reconstruction makes an approximate estimate of the input, seen in \subref{fig:seq_pca}. The zero level set of the residual is shown in \subref{fig:seq_res}.}
\label{fig:sequential_problem}
\vspace{-0.65cm}
\end{center}
\end{figure}
\begin{figure}[h]
\begin{centering}
\includegraphics[width =0.8\linewidth]{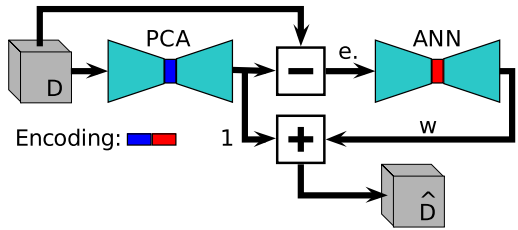}
\caption{Coupling the PCA and ANN encoder/decoder pairs in a sequential manner. The ANN part is trained on a data-set that consists of the residuals of the corresponding PCA encoder/decoder reconstruction relative to the original data-set.}
\label{fig:sequential}
\end{centering}
\end{figure}

The second extension we propose is based on the observation that the difference between the decoded data and the input still contains a lot of low-frequency variation c.f. Fig.\ref{fig:sequential_problem}, even if it is increasingly complex and non-linear. In the limit of what can be achieved, it would be expected that the residual should converge to a random signal. Being far from this, however,  we may attempt to model the residual and add it to the result of the first stage decoding as shown in Fig. \ref{fig:sequential}. 
This entails that for each different first-stage component, a new data-set must be generated, containing the residuals relative to the original TSDF data. The second stage is then trained to model these residuals instead of the original data.
During encoding, the TSDF is passed to the first stage (in this case PCA). The data is encoded and decoded by the first stage and the decoded result is subtracted from the original input. The resulting residual is encoded by the second stage and their code vectors are concatenated.
For decoding, each stage processes their respective codes independently, and the result is added with a weight applied only to the second stage output (which now contains both negative and positive values). Since the residuals are centred around zero, we choose to use the hyperbolic tangent as activation function for the sequential ANN decoder.
To study the effects of the various algorithmic choices, the code (or feature) size is limited to \textit{128} floating point values. When using mixed encoding strategies, the dimensions for each component are therefore chosen to be complementary (totaling \textit{128}). Since the PCA encoder/decoders were designed with compact representations of \textit{32} and \textit{64}  elements, we train the ANN's with code-layers of \textit{96} and \textit{64}  elements, respectively.
%

\subsection{Evaluation Methodology}
\label{sec:evaluation}

Given the fixed-sized feature or code-vector, how do we best allocate its elements? And which combination method is best? We explore these question by means of two different fitness quality measures. Reconstruction fidelity and ego-motion estimation. 
To aid in our analysis we use a publicly available RGB-D data-set \cite{sturm12iros} with ground-truth pose estimates provided by an independent external camera-tracking system. Using the provided ground-truth poses, we generate a map, by fusing the depth images into a TSDF representation. This produces a ground truth map. We chose \textit{teddy, room, desk, desk2, 360} and \textit{plant} from the \textit{freiburg-1} collection for evaluation as these are representative of real-world challenges that arise in SLAM and visual odometry, including motion blur, sensor noise and occasional lack of geometric structure needed for tracking. We do not use the RGB components of the data for any purpose in this work.

\subsubsection{Reconstruction Error}
As a measure for reconstruction error, we compute the mean squared errors of the decoded distance fields relative to the input. This metric is relevant to path planning, manipulation and object detection tasks since it indirectly relates to the fidelity of surface locations. For each data-set, using each encoder/decoder we compute a lossy version of the original data and report the average and standard deviation across all data-sets.

\subsubsection{Ego-motion Estimation}
Ego-motion estimation performance is measured by the absolute trajectory error (ATE)\cite{sturm12iros}. The absolute trajectory error is the integrated distance between all pose estimates relative to the ground truth trajectory. The evaluations are performed by loading a complete TSDF map into memory and setting the initial pose according to ground truth. Then, as depth images are loaded from the RGB-D data-set, we estimate the camera transformation that minimizes the point to model distance for each new frame. The evaluation was performed on all the data-sets, processed through each compression and subsequent decompression method. As a baseline, we also included the original map, processed with a Gaussian blur kernel of size 9x9x9 voxels and a $\sigma$ parameter of $4/3$.

\subsubsection{Implementation Notes}
The PCA basis was produced, using the dimensionality reduction tools from the \textbf{scikit-learn} \cite{scikit-learn} library. Autoencoders were trained using \textbf{pylearn2} \cite{goodfellow2013pylearn2} using batch gradient descent with the change in reconstruction error on a validation data-set as a stopping criterion. The data-set was split into $400$ batches containing $500$ samples each, of which $300$ batches were used for training, $50$ for testing, and $50$ for validation. The networks use \textit{sigmoid} activation units and contain $4096, 512, d, 512, 4096$ nodes with $d$ representing the number of dimensions of the descriptor. 

The runtime implementation for all the encoder/decoder architectures was done using cuBLAS\footnote{https://developer.nvidia.com/cuBLAS} and Thrust\footnote{https://developer.nvidia.com/Thrust} libraries for GPU-enabled matrix-vector and array computation. 
Timing the execution of copying data to the GPU, encoding, decoding and copying it back to main memory gives an average time of $405 - 645 \mu s$ per block of $16^3$ voxels. This is likely to be a conservative run-time estimate for practical scenarios since the memory transfers, which represents the major part of the time, would most likely be made in feature space (and in batches) rather than block by block in the voxel domain. Furthermore, only one of the operations (compression or decompression) would typically be required, not both. The span in timing depends on the encoding method used, sequential encoding representing the upper bound and PCA-based encoding, the lower.
\section{Experimental Results}\label{sec:results}
\subsection{Reconstruction Error}
\begin{figure}[t]
\begin{center}
\subfigure[] {
        \includegraphics[width =.45\linewidth]{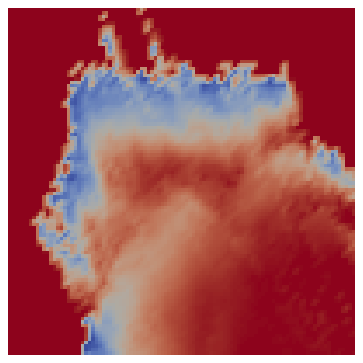}
	    \label{fig:field_original}
}
\subfigure[] {
        \includegraphics[width =.45\linewidth]{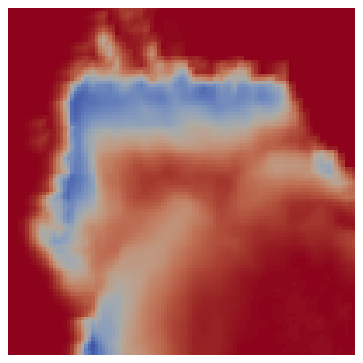}	
        \label{fig:field_gaussian}
}
\subfigure[] {
        \includegraphics[width =.45\linewidth]{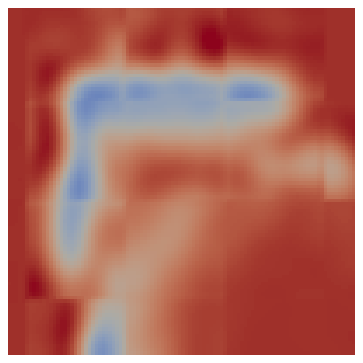}
   	    \label{fig:field_pca}
}
\subfigure[] {
        \includegraphics[width =.45\linewidth]{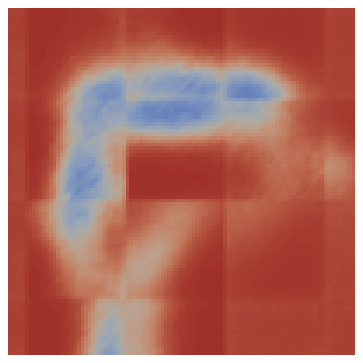}
	    \label{fig:field_nn}
}
\vspace{-0.25cm}
\caption{A slice through the distance field reconstructed through different methods, using 64-element encodings. Shown here are \subref{fig:field_original} the original map, \subref{fig:field_gaussian} the Gaussian filtered map, \subref{fig:field_pca} PCA reconstruction and \subref{fig:field_nn} auto-encoder reconstruction }
 \label{fig:field_compare}
\vspace{-0.65cm}
\end{center}
\end{figure}
We report the average reconstruction error over all non-empty blocks in all data-sets and the standard deviation among data-sets in Table \ref{table:results}. The reconstruction errors obtained strongly suggest that increasing the size of the codes for individual encoders yields better performance, though with diminishing returns. Several attempts were made, to out-perform the PCA approach, using Artificial Neural Networks (ANN) trained as auto-encoders but this was generally unsuccessful.
PCA-based encoders, using \textit{32, 64} and \textit{128} components, produce better results than ANN encoders in all our experiments.  We also noted that when searching for the optimal mixing weight for the parallel and sequential encoding architectures, mixing is rarely advantageous. For the parallel method it is most often preferable to choose one encoder or the other (most often PCA), effectively wasting half of the encoding space. In the sequential method, it is most often best not to include the ANN at all, or with near-zero weight. We include only the results where we employed a \textit{64-64} component split and note from other experiments that these architectures generally perform on par with the PCA-only solution of respective dimensionality e.g. 64 in the reported case.
\def\arraystretch{1.25}%
\begin{table*}[ht]
\begin{centering}
    \begin{tabular}{|l|c|c|c|}
  \hline
    Reconstruction Method   & Reconstruction Error (MSE)$\pm\sigma$ & Mean ATE [m] $\pm\sigma$          & Median ATE [m]\\ \hline
    Original data           & -                                     & 0.70$\pm$ 0.67                    & 0.59          \\ \hline
    PCA 32                  & 42.94 $\pm$ 2.63                      & \textbf{0.29} $\pm$\textbf{0.45}  & \textbf{0.06} \\ \hline
    PCA 64                  & 33.96 $\pm$ 2.01                      & 0.48 $\pm$ 0.53                   & 0.16          \\ \hline
    PCA 128                 &\textbf{27.29} $\pm$\textbf{1.87}      & 0.65 $\pm$ 0.54                   & 0.62          \\ \hline
    NN 32                   & 59.65 $\pm$ 2.78                      & 0.093 $\pm$ 0.11                  & 0.07          \\ \hline
    NN 64                   & 49.52 $\pm$ 2.19                      &\textbf{0.083} $\pm$\textbf{0.10}  & 0.06          \\ \hline
    NN 128                  & 46.19 $\pm$ 2.23                      & 0.087 $\pm$ 0.12                  &\textbf{0.05}  \\ \hline
    Parallel PCA 64+NN 64   & 33.63 $\pm$ 1.98                      & 0.27 $\pm$ 0.39                   & 0.07          \\ \hline
    Sequential PCA 64+NN 64 & 33.95 $\pm$ 2.01                      & 0.49 $\pm$ 0.56                   & 0.16          \\ \hline
    Gaussian Blur 9x9x9     &  -                                    & \textbf{0.05}$\pm$\textbf{0.04}   &\textbf{0.04}  \\ \hline
    \end{tabular}
    \caption {Average reconstruction and ego-motion estimation results across all data-sets.}
    \label{table:results}
    \end{centering}
\vspace{-1.0cm}
\end{table*}
\par
\begin{figure}[t]
\begin{center}
\subfigure[] {
        \includegraphics[width =.75\linewidth]{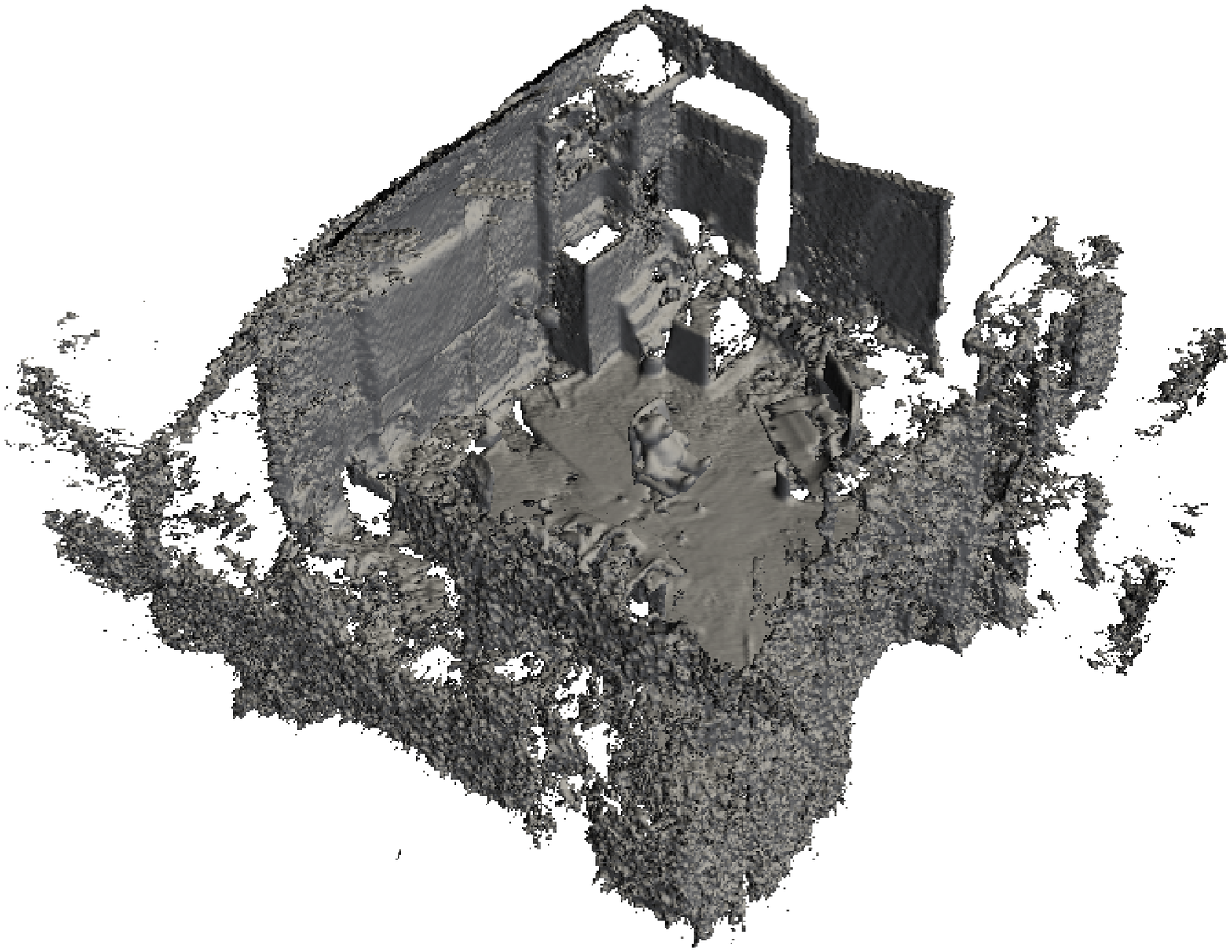}
}
\subfigure[] {
        \includegraphics[width =.75\linewidth]{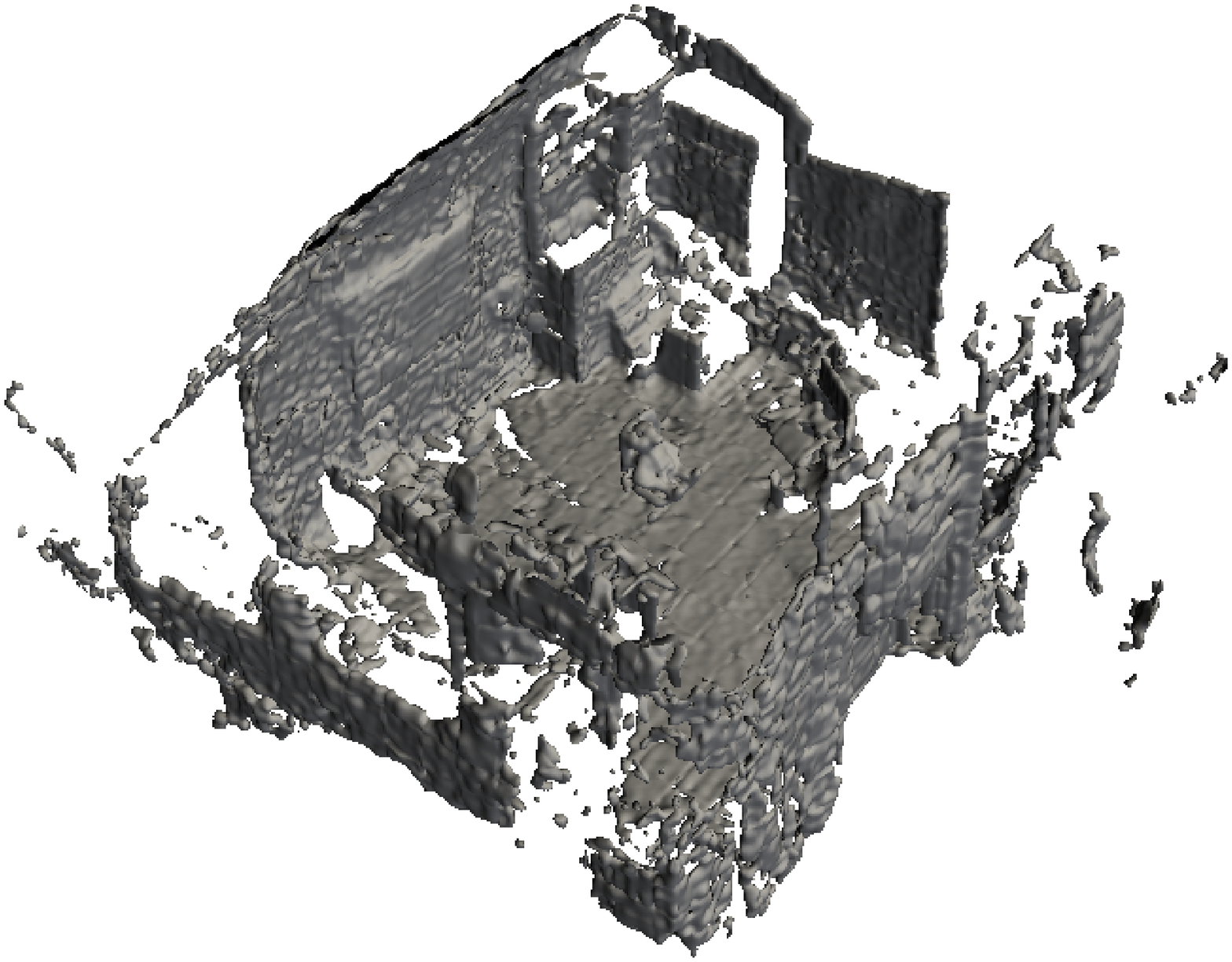}
        \label{fig:teddy_reconst:b}
}
\vspace{-0.25cm}
\caption{Example reconstruction using a PCA basis with 128 components. The reconstructed version \subref{fig:teddy_reconst:b} includes some blocking artifacts, visible as tiles on the floor of the room, but contains visibly less noise.}
 \label{fig:teddy_reconst}
\vspace{-0.65cm}
\end{center}
\end{figure}
The best overall reconstruction performance is given by the baseline PCA encoder/decoder, using 128 components. We illustrate this with an image from the \textit{teddy} data-set, in Fig. \ref{fig:teddy_reconst}. Note that the decoded data-set is smoother, so in a sense the measured discrepancy is partly related to a qualitative improvement.
\subsection{Ego-motion Estimation}
The ego-motion estimation, performed by the SDF Tracker algorithm, uses the TSDF as a cost function to which subsequent 3D points are aligned. This requires that the gradient of the TSDF be of correct magnitude and point in the right direction. To get a good alignment, the minimum absolute distance should coincide with the actual location of the surface. 
\par
In spite of being given challenging camera trajectories, performance using the decoded maps is on average better than the unaltered map. When the tracker keeps up with the camera motion, we have observed that the performance resulting from the use of each map is in the order of their respective reconstruction errors. In this case, the closer the surface is to the ground truth model, the better. 
However tracking may fail for various reasons, e.g. when there is little overlap between successive frames, when the model or depth image contains noise or when there is not enough geometric variation to properly constrain the pose estimation. In some of these cases, the maps that offer simplified approximations to the original distance field fare better. The robustness in tracking is most likely owed to the denoising effect that the encoding has, as evidenced by the performance on the Gaussian blurred map. Of the encoded maps, we see that the AE compression results in better pose estimation. In Fig. \ref{fig:field_compare} we see a slice through a volume colour-coded by distance. Here we note that even though the PCA-based map is more similar to the original, on the left side of the image it is evident that the field is not monotonically increasing away from the surface. Such artefacts cause the field gradient to point in the wrong direction, possibly contributing to failure to find the correct alignment.
The large difference between the median and mean values for the pose estimation errors are indicative of mostly accurate pose estimations, with occasional gross misalignments. 
\subsection{Selective Feature-based Map Expansion}
\begin{figure}[t]
\begin{center}
\subfigure[] {
        \includegraphics[width =.8\linewidth]{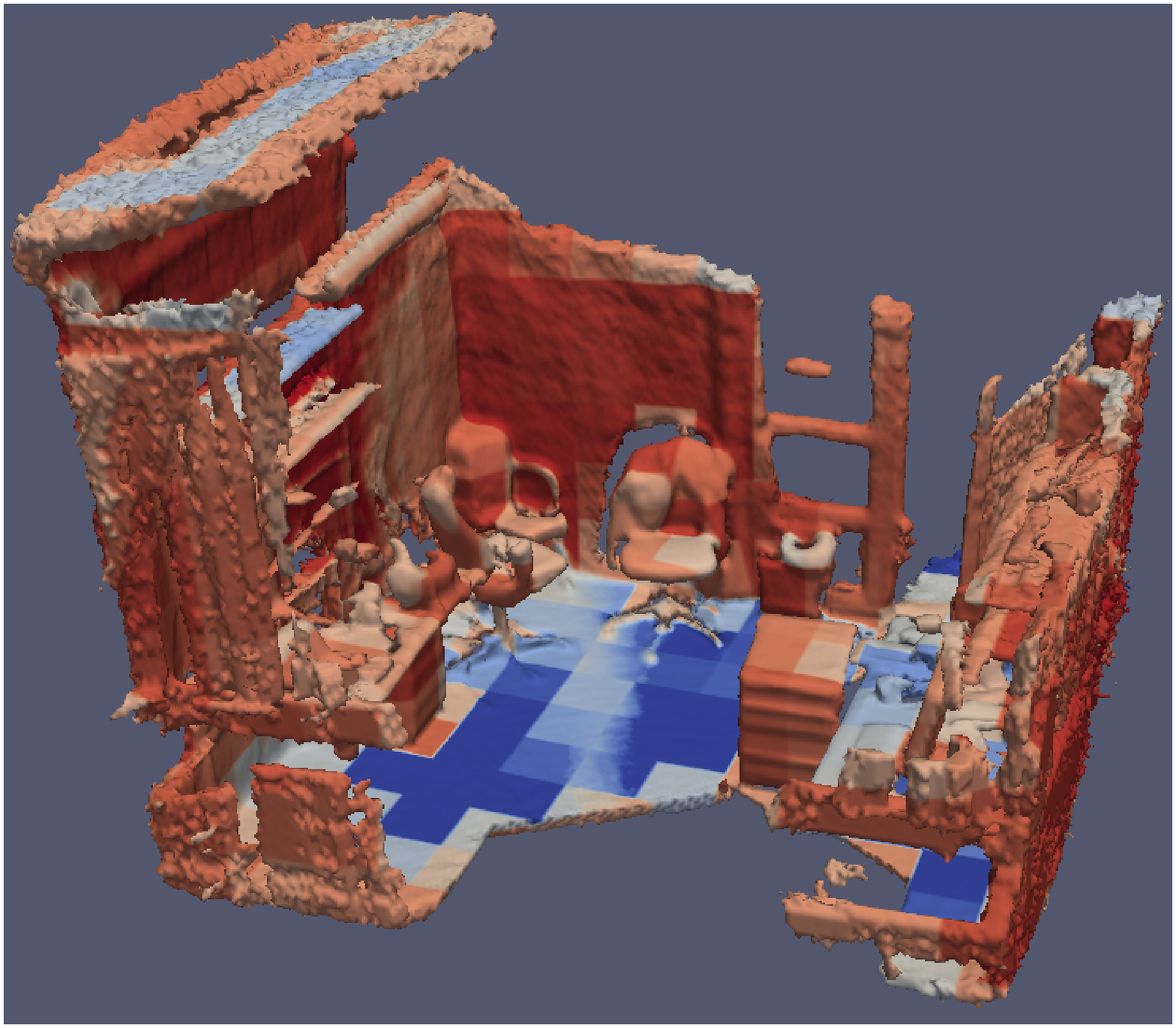}
}
\subfigure[] {
        \includegraphics[width =.8\linewidth]{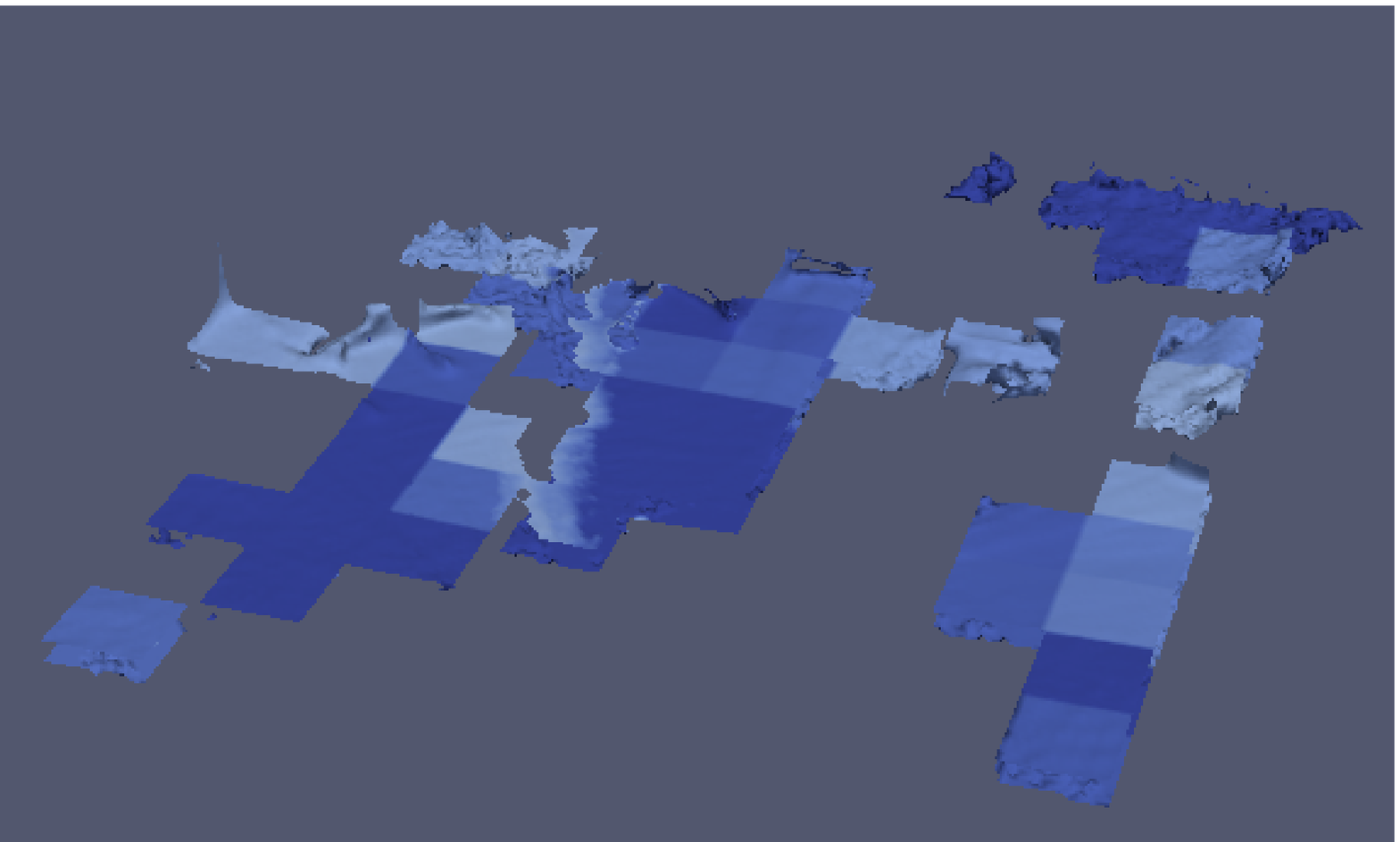}
}
\vspace{-0.25cm}
\caption{Selective reconstruction of floor surfaces. Given a compressed map, the minimum distance for each compressed block, to a set of descriptors that relate to horizontal planes can be computed (e.g. floors). Only the blocks that are similar enough to this set of descriptors need to be considered for actual decompression. In the first figure, the uncompressed map is shown, with each region coloured according to its descriptor's distance to the set of descriptors that relate to floors. In the second figure, we see the selectively expanded floor cells.}
 \label{fig:floor_segmentation}
\vspace{-0.65cm}
\end{center}
\end{figure}
Although the descriptors we obtain are clearly not invariant to affine transformations (if they were, the decompression wouldn't reproduce the field in its correct location/orientation), we can still create descriptor-based models for geometries of particular interest by sampling their TSDFs over the range of transformations to which we want the model to be invariant. If information about the orientation of the map is known a priori, e.g. some dominant structures are axis-aligned with the voxel lattice, or dominant structures are orthogonal to each other, the models can be made even smaller. 
In the example illustrated in Fig. \ref{fig:floor_segmentation}, a descriptor-based model for floors was first created by encoding the TSDFs of horizontal planes at 15 different offsets, generating one 64-element vector each. Each descriptor in the compressed map can then be compared to this small model by the squared norm of their differences and only those beneath a threshold of similarity need to be considered for expansion. 
Here an advantage of the PCA-based encoding becomes evident: Since PCA generates its linear subspace in an ordered manner, feature vectors of different dimensionality can be tested for similarity up to the number of elements of the smallest, i.e., a 32-dimensional feature descriptor can be matched against the first half of a 64-dimensional feature descriptor. This property is useful in handling multiple levels of compression, for different applications, whilst maintaining a common way to describe them.
\section{Conclusions}\label{sec:conclusions}
In this paper, we presented the use of dimensionality reduction of TSDF volumes, which lie at the core of many algorithms across a wide domain of applications with close ties to robotics. We proposed PCA and ANN encoding strategies as well as hybrid methods and evaluated their performance with respect to a camera tracking application and to reconstruction error.

We demonstrate that we can compress volumetric data using PCA and neural nets to small sizes (between 128:1 and 32:1) and still use them in camera tracking applications with good results. We show that PCA produces superior reconstruction results and although neural nets have inherently greater expressive power, training them is not straightforward, often resulting in lower quality reconstructions but nonetheless offering slightly better performance in ego-motion estimation applications.
We found that combining encoders in parallel with optimal mixture weights usually leads to Either/Or situations, and more seldom using both simultaneously. The sequential combination of encoders is rarely an advantage, possibly due to the residual being an overly complex function to model. 
Finally, we have shown that this entire class of methods can be successfully applied to both compress and imbue the data with some low-level semantic meaning and suggested an application in which both of these characteristics are simultaneously desirable.
\section{Future Work}\label{sec:future}
It is clear that the resulting features are not invariant to rigid-body transformations and experimentally matching features of identical objects in different poses, suggests that features do not form object-centred clusters in the lower-dimensional space. A method for obtaining a low-dimensional representation as well as a reliable transformation into some canonical frame of reference would pave the way for many interesting applications in semantic mapping and scene understanding. 
Furthermore, it seems unfortunate that pose-estimation ultimately has to occur in the voxel domain. Given that the transformation to the low dimensional space is a simple affine function (at least for the PCA-based encoding) it seems intuitive that one should be able to formulate and solve the pose-estimation problem in the reduced space with a lower memory requirement in all stages of computation. Investigating this possibility remains an interesting problem as it is not clear if this would represent a direct trade-off between memory complexity and computational complexity.
%
%
\section*{Acknowledgement}
This work has partly been supported by the European Commission under contract number FP7-ICT-270350 (RobLog).
\addtolength{\textheight}{-5.0cm}
%



%
%


\bibliographystyle{styles/IEEEtran}
\bibliography{references}

\begin{thebibliography}{10}
\providecommand{\url}[1]{#1}
\csname url@rmstyle\endcsname
\providecommand{\newblock}{\relax}
\providecommand{\bibinfo}[2]{#2}
\providecommand\BIBentrySTDinterwordspacing{\spaceskip=0pt\relax}
\providecommand\BIBentryALTinterwordstretchfactor{4}
\providecommand\BIBentryALTinterwordspacing{\spaceskip=\fontdimen2\font plus
\BIBentryALTinterwordstretchfactor\fontdimen3\font minus
  \fontdimen4\font\relax}
\providecommand\BIBforeignlanguage[2]{{%
\expandafter\ifx\csname l@#1\endcsname\relax
\typeout{** WARNING: IEEEtran.bst: No hyphenation pattern has been}%
\typeout{** loaded for the language `#1'. Using the pattern for}%
\typeout{** the default language instead.}%
\else
\language=\csname l@#1\endcsname
\fi
#2}}

\bibitem{fitzgibbon2003robust}
A.~W. Fitzgibbon, ``{Robust Registration of 2D and 3D Point Sets},''
  \emph{Image and Vision Computing}, vol.~21, no.~13, pp. 1145--1153, 2003.

\bibitem{hart1996sphere}
J.~C. Hart, ``Sphere tracing: A geometric method for the antialiased ray
  tracing of implicit surfaces,'' \emph{The Visual Computer}, vol.~12, no.~10,
  pp. 527--545, 1996.

\bibitem{fuhrmann2003distance}
A.~Fuhrmann, G.~Sobotka, and C.~Gro{\ss}, ``Distance fields for rapid collision
  detection in physically based modeling,'' in \emph{Proceedings of GraphiCon
  2003}, 2003, pp. 58--65.

\bibitem{hoff1999fast}
K.~E. Hoff~III, J.~Keyser, M.~Lin, D.~Manocha, and T.~Culver, ``Fast
  computation of generalized voronoi diagrams using graphics hardware,'' in
  \emph{Proceedings of the 26th annual conference on Computer graphics and
  interactive techniques}.\hskip 1em plus 0.5em minus 0.4em\relax ACM
  Press/Addison-Wesley Publishing Co., 1999, pp. 277--286.

\bibitem{schmidt2014dart}
T.~Schmidt, R.~Newcombe, and D.~Fox, ``{DART}: Dense articulated real-time
  tracking,'' in \emph{Proceedings of Robotics: Science and Systems}, Berkeley,
  USA, July 2014.

\bibitem{curless1996volumetric}
B.~Curless and M.~Levoy, ``A volumetric method for building complex models from
  range images,'' in \emph{Proceedings of the 23rd annual conference on
  Computer graphics and interactive techniques}.\hskip 1em plus 0.5em minus
  0.4em\relax ACM, 1996, pp. 303--312.

\bibitem{newcombe2011kinectfusion}
R.~A. Newcombe, A.~J. Davison, S.~Izadi, P.~Kohli, O.~Hilliges, J.~Shotton,
  D.~Molyneaux, S.~Hodges, D.~Kim, and A.~Fitzgibbon, ``{KinectFusion:
  Real-time dense surface mapping and tracking},'' in \emph{Mixed and Augmented
  Reality (ISMAR), 2011 10th IEEE International Symposium on}, 2011, pp.
  127--136.

\bibitem{whelan2012kintinuous}
T.~Whelan, M.~Kaess, M.~Fallon, H.~Johannsson, J.~Leonard, and J.~McDonald,
  ``Kintinuous: Spatially extended {K}inect{F}usion,'' in \emph{RSS Workshop on
  RGB-D: Advanced Reasoning with Depth Cameras}, Sydney, Australia, Jul 2012.

\bibitem{roth2012moving}
H.~Roth and M.~Vona, ``Moving volume kinectfusion.'' in \emph{BMVC}, 2012, pp.
  1--11.

\bibitem{canelhas2013sdf}
D.~R. Canelhas, T.~Stoyanov, and A.~J. Lilienthal, ``Sdf tracker: A parallel
  algorithm for on-line pose estimation and scene reconstruction from depth
  images,'' in \emph{Intelligent Robots and Systems (IROS), 2013 IEEE/RSJ
  International Conference on}.\hskip 1em plus 0.5em minus 0.4em\relax IEEE,
  2013, pp. 3671--3676.

\bibitem{bylow2013sdf}
E.~Bylow, J.~Sturm, C.~Kerl, F.~Kahl, and D.~Cremers, ``Real-time camera
  tracking and 3d reconstruction using signed distance functions,'' in
  \emph{Proceedings of Robotics: Science and Systems}, Berlin, Germany, June
  2013.

\bibitem{elfes1989occupancy}
A.~Elfes, ``Occupancy grids: A probabilistic framework for robot perception and
  navigation,'' 1989.

\bibitem{frisken2000adaptively}
S.~F. Frisken, R.~N. Perry, A.~P. Rockwood, and T.~R. Jones, ``Adaptively
  sampled distance fields: a general representation of shape for computer
  graphics,'' in \emph{Proceedings of the 27th annual conference on Computer
  graphics and interactive techniques}.\hskip 1em plus 0.5em minus 0.4em\relax
  ACM Press/Addison-Wesley Publishing Co., 2000, pp. 249--254.

\bibitem{zeng2012memory}
M.~Zeng, F.~Zhao, J.~Zheng, and X.~Liu, ``A memory-efficient kinectfusion using
  octree,'' in \emph{Computational Visual Media}.\hskip 1em plus 0.5em minus
  0.4em\relax Springer, 2012, pp. 234--241.

\bibitem{newcombe2014phd}
R.~A. Newcombe, ``{Dense Visual SLAM},'' Ph.D. dissertation, Imperial College
  London, United Kingdom, June 2014.

\bibitem{wold1987principal}
S.~Wold, K.~Esbensen, and P.~Geladi, ``Principal component analysis,''
  \emph{Chemometrics and intelligent laboratory systems}, vol.~2, no.~1, pp.
  37--52, 1987.

\bibitem{ruhnke2013compact}
M.~Ruhnke, L.~Bo, D.~Fox, and W.~Burgard, ``Compact rgbd surface models based
  on sparse coding.'' in \emph{AAAI}, 2013.

\bibitem{aharon2006svd}
M.~Aharon, M.~Elad, and A.~Bruckstein, ``-svd: An algorithm for designing
  overcomplete dictionaries for sparse representation,'' \emph{Signal
  Processing, IEEE Transactions on}, vol.~54, no.~11, pp. 4311--4322, 2006.

\bibitem{mao2014active}
\BIBentryALTinterwordspacing
J.~Mao, J.~Zhu, and A.~Yuille, ``\BIBforeignlanguage{English}{An active patch
  model for real world texture and appearance classification},'' in
  \emph{\BIBforeignlanguage{English}{Computer Vision – ECCV 2014}}, ser.
  Lecture Notes in Computer Science, D.~Fleet, T.~Pajdla, B.~Schiele, and
  T.~Tuytelaars, Eds.\hskip 1em plus 0.5em minus 0.4em\relax Springer
  International Publishing, 2014, vol. 8691, pp. 140--155. [Online]. Available:
  \url{http://dx.doi.org/10.1007/978-3-319-10578-9_10}
\BIBentrySTDinterwordspacing

\bibitem{turk1991eigenfaces}
M.~Turk and A.~Pentland, ``Eigenfaces for recognition,'' \emph{Journal of
  cognitive neuroscience}, vol.~3, no.~1, pp. 71--86, 1991.

\bibitem{richardson2004h}
I.~E. Richardson, \emph{H. 264 and MPEG-4 video compression: video coding for
  next-generation multimedia}.\hskip 1em plus 0.5em minus 0.4em\relax John
  Wiley \& Sons, 2004.

\bibitem{marcellin2002jpeg2000}
M.~W. Marcellin, \emph{JPEG2000 Image Compression Fundamentals, Standards and
  Practice: Image Compression Fundamentals, Standards, and Practice}.\hskip 1em
  plus 0.5em minus 0.4em\relax springer, 2002, vol.~1.

\bibitem{jones2004distance}
M.~W. Jones, ``Distance field compression,'' 2004.

\bibitem{rumelhart1988learning}
D.~E. Rumelhart, G.~E. Hinton, and R.~J. Williams, ``Learning representations
  by back-propagating errors,'' \emph{Cognitive modeling}, 1988.

\bibitem{hinton2006reducing}
G.~E. Hinton and R.~R. Salakhutdinov, ``Reducing the dimensionality of data
  with neural networks,'' \emph{Science}, vol. 313, no. 5786, pp. 504--507,
  2006.

\bibitem{distfun}
\BIBentryALTinterwordspacing
I.~Qu\`ilez, ``Modeling with distance functions,'' 2008. [Online]. Available:
  \url{https://www.iquilezles.org/www/articles/distfunctions/distfunctions.htm}
\BIBentrySTDinterwordspacing

\bibitem{dietterich2000ensemble}
T.~G. Dietterich, ``Ensemble methods in machine learning,'' in \emph{Multiple
  classifier systems}.\hskip 1em plus 0.5em minus 0.4em\relax Springer, 2000,
  pp. 1--15.

\bibitem{sturm12iros}
J.~Sturm, N.~Engelhard, F.~Endres, W.~Burgard, and D.~Cremers, ``A benchmark
  for the evaluation of rgb-d slam systems,'' in \emph{Proc. of the
  International Conference on Intelligent Robot Systems (IROS)}, Oct. 2012.

\bibitem{scikit-learn}
F.~Pedregosa, G.~Varoquaux, A.~Gramfort, V.~Michel, B.~Thirion, O.~Grisel,
  M.~Blondel, P.~Prettenhofer, R.~Weiss, V.~Dubourg, J.~Vanderplas, A.~Passos,
  D.~Cournapeau, M.~Brucher, M.~Perrot, and E.~Duchesnay, ``Scikit-learn:
  Machine learning in {P}ython,'' \emph{Journal of Machine Learning Research},
  vol.~12, pp. 2825--2830, 2011.

\bibitem{goodfellow2013pylearn2}
I.~J. Goodfellow, D.~Warde-Farley, P.~Lamblin, V.~Dumoulin, M.~Mirza,
  R.~Pascanu, J.~Bergstra, F.~Bastien, and Y.~Bengio, ``Pylearn2: a machine
  learning research library,'' \emph{arXiv preprint arXiv:1308.4214}, 2013.

\end{thebibliography}
\end{document}